\documentclass[11pt]{article}

\usepackage[margin=1in]{geometry}
\usepackage{amsmath,amssymb,amsthm}
\usepackage{mathtools}
\usepackage{graphicx}
\usepackage{booktabs}
\usepackage{microtype}
\usepackage[hidelinks]{hyperref}
\usepackage{url}

\newtheorem{proposition}{Proposition}

\newtheorem{remark}{Remark}

\DeclareMathOperator{\col}{col}

\DeclareMathOperator{\softmax}{softmax}
\newcommand{\R}{\mathbb{R}}
\newcommand{\W}{\mathbf{W}}
\newcommand{\hidden}{g}
\newcommand{\Ahat}{\hat{A}}

\title{\textbf{The Geometry of Last-Layer Model Stealing}}

\author{%
  Snigdha Chandan Khilar\thanks{Independent Researcher.}\\
  \texttt{snkhilar@gmail.com}
}
\date{\today}

\begin{document}
\maketitle

\begin{abstract}
We give a geometric reading of the last-layer model-stealing attack of
Carlini et al.~[\ref{ref:carlini}], using the vocabulary of exterior differential
systems (EDS) recently extended to Lie algebroids by Hohloch, Mestdag and
Yasaka~[\ref{ref:eds}]. The set of logit vectors a transformer can emit is the
common zero locus of an \emph{ideal} with one linear part (recovered by the
singular value decomposition) and one quadratic part (the ellipsoid induced by
the final normalization layer). In this language the object that controls
recovery of the projection matrix is the \emph{polar space} of that quadric,
which we show is exactly the tangent space of the output manifold; recovery
succeeds, up to the unavoidable orthogonal gauge, precisely when a pair of
\emph{regularity} conditions---the analogues of K\"ahler-regularity---hold. We
verify every step on a fully controlled toy model to machine precision. We then
ask what lies below the last layer and report two things. First, the
\emph{intrinsic dimension} of the recoverable hidden-state manifold is an
observable, invisible to the singular value decomposition and to the quadric,
that detects a nonlinear sublayer and measures its effective rank. Second, we
characterize what is and is not identifiable beneath the last layer, and exhibit
large explicit non-identifiable fibers: different sublayers, and even different
architecture widths, that produce bit-identical outputs. We are deliberately
explicit about scope: the EDS framing organizes the picture but is not the
engine, and the load-bearing results are classical. The contribution is a clean
unified account and a concrete identifiability boundary, not a new attack.
\end{abstract}

\section{Introduction}

Production language models are exposed only through APIs, yet
Carlini et al.~[\ref{ref:carlini}] showed that the final
\emph{embedding projection} (``unembedding'') layer of such a model can be
recovered, up to symmetries, from ordinary query access. Their attack is
top-down: because the last layer maps a small hidden dimension $h$ to a large
logit vector of dimension $l\gg h$, the logits live in an $h$-dimensional
subspace, and the singular value decomposition (SVD) of enough query responses
reveals both $h$ and the projection matrix up to a linear change of basis. A
refinement (their Appendix~H) exploits the fact that the final normalization
layer places hidden states on a sphere, so the logits lie on an ellipsoid;
fitting that ellipsoid sharpens the recovery from ``up to an invertible matrix''
to ``up to an orthogonal matrix.''

This note makes two contributions, both modest and clearly delimited.

\paragraph{A geometric account.} We recast the attack in the language of
exterior differential systems, which is the natural setting for ``reconstruct a
global object from local data under constraints, modulo a symmetry group.'' The
attainable logits form the integral variety of an \emph{ideal}; the object
governing recovery of the projection is the \emph{polar space} of its quadratic
generator; and recovery is well posed exactly under regularity conditions that
are the affine analogues of the K\"ahler-regularity used in the Cartan--K\"ahler
theorem~[\ref{ref:eds},\ref{ref:bryant}]. The single-layer case turns out to be
Frobenius-integrable---which is precisely why the attack is closed-form rather
than iterative---and we confirm the whole picture numerically.

\paragraph{An identifiability wall.} We then look one layer deeper. We observe
that the \emph{intrinsic dimension} of the recoverable hidden-state manifold is
an extraction observable distinct from the linear span the SVD measures: when a
low-rank nonlinear sublayer is present, the span overstates the content
dimension, and the intrinsic dimension reveals the bottleneck. Finally we give a
crisp identifiability characterization of the sublayer and demonstrate, with
machine-precision examples, that most of its parameters lie in a non-identifiable
fiber. This explains mechanically why the attack has not been extended past one
layer: it is not a missing trick but a property of the observation map.

\paragraph{Honesty about scope.} The geometric language is organizing, not
enabling: at no point does it produce a result the standard linear-algebra and
manifold tools could not. The identifiability statements rest on classical
neural-network identifiability~[\ref{ref:sussmann}] and on the known fact that
learned representations have low intrinsic dimension~[\ref{ref:ansuini}]. We
state this plainly so the note is read as a unified exposition with a concrete
identifiability boundary, not as a new attack.

\paragraph{Roadmap.} Section~\ref{sec:background} is a self-contained primer on
both halves of the story---model stealing and the handful of
differential-geometry notions we borrow---written for readers who know neither;
specialists can skip to Section~\ref{sec:setup}. Sections~\ref{sec:ideal}--\ref{sec:reg}
develop and verify the single-layer picture, and
Sections~\ref{sec:below}--\ref{sec:wall} look beneath the last layer.

\section{Background and intuition}\label{sec:background}

This section assumes no prior exposure to either model extraction or exterior
differential systems (EDS). Experts may skip to Section~\ref{sec:setup}.

\paragraph{What ``model stealing'' means.} A language model is usually served
behind an API: you send text and receive, for each possible next token, a score
(a \emph{logit}) that the model turns into a probability. The provider keeps the
model's weights secret. \emph{Model stealing} asks how much of those weights an
outsider can reconstruct using only API queries. One does not expect to copy a
multi-billion-parameter model from query access; the surprising result
of~[\ref{ref:carlini}] is that one specific piece---the final linear layer---can
be recovered exactly, up to an unavoidable ambiguity, and cheaply.

\paragraph{Why the last layer is the easy target.} A transformer carries
information in a vector of width $h$ (the ``hidden'' or ``residual'' dimension),
then multiplies that vector by a matrix $\W$ to produce one logit per
vocabulary token. The vocabulary is large ($l$ in the tens of thousands) while
$h$ is comparatively small, so $\W$ is a tall, thin, rank-$h$ matrix: it maps a
small space up into a large one. That gap, $h\ll l$, is the crack the attack
pries open.

\begin{figure}[t]
  \centering
  \includegraphics[width=0.62\linewidth]{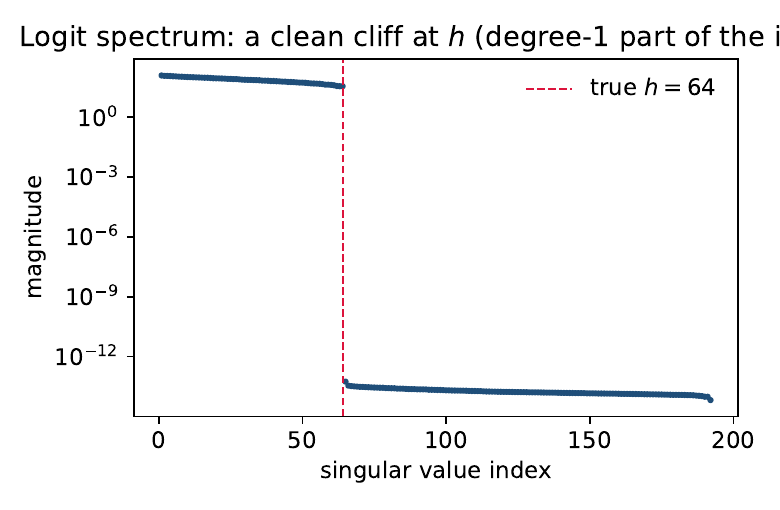}
  \caption{The degree-1 part of the ideal. On a toy model with $h=64$, the
  logit singular spectrum drops by fourteen orders of magnitude at index $h$:
  the recovered $\hat h$ is exactly $64$.}
  \label{fig:spectrum}
\end{figure}

\paragraph{The rank trick.} Query the model on many different prompts and stack
the logit vectors as columns of a matrix. Although each column lives in $l$
dimensions, every column is $\W$ times something $h$-dimensional, so all of them
lie in the same $h$-dimensional subspace. Once you have queried more than $h$
times, new responses become linear combinations of old ones. The singular value
decomposition (SVD) detects this: it reports exactly $h$ large singular values
and a sharp drop afterwards. Counting them recovers the hidden width
(Figure~\ref{fig:spectrum}); a little more linear algebra recovers $\W$ itself,
up to a change of basis.

\paragraph{From a sphere to an ellipsoid.} Modern transformers
\emph{normalize} the hidden vector just before the last layer, which forces it
to have fixed length---it lives on a sphere. A linear map sends a sphere to an
\emph{ellipsoid}. So the logits do not merely fill an $h$-dimensional subspace;
they lie on an ellipsoidal surface inside it. Fitting that ellipsoid pins down
more of $\W$: it sharpens ``known up to any invertible change of basis'' to
``known up to a rotation.'' The leftover rotation is genuinely unrecoverable, for
a simple reason given in Section~\ref{sec:polar}.

\paragraph{Three borrowed ideas.} The EDS vocabulary we use names three things
that are already implicitly present above.
\begin{itemize}
\item An \emph{ideal} is just the collection of equations every observation
satisfies. Here there are two kinds: linear ones (the logits lie in the
subspace) and one quadratic one (they lie on the ellipsoid). The surface they
cut out is the \emph{integral variety}---the set of attainable outputs.
\item A \emph{polar space} answers ``given part of a solution, which directions
can extend it?'' For a quadric this is the classical notion of points
\emph{conjugate} with respect to the surface, and---as we show---it is exactly
the tangent plane to the ellipsoid. Recovering the last layer amounts to
recovering this field of tangent planes.
\item \emph{Regularity} is the package of nondegeneracy conditions that make the
reconstruction unique and stable: a clean gap in the spectrum, and a genuinely
curved (nondegenerate) ellipsoid. When they fail, the attack fails in a
predictable way.
\end{itemize}

\paragraph{Why bring in EDS at all?} Honestly, for the single layer it is a
unifying language rather than a new tool: rank recovery, ellipsoid recovery, the
rotation ambiguity, and the stability conditions become one object with three
features. Its real payoff is conceptual---it tells us in advance
(Remark~\ref{rem:frobenius}) why the single-layer attack is one-shot, and it
frames the genuinely hard question, ``what can be learned about the layer
underneath,'' as a question about the geometry of a curved surface
(Sections~\ref{sec:below}--\ref{sec:wall}).

\section{Setup}\label{sec:setup}

Let $\mathcal{X}$ be the token vocabulary, $|\mathcal{X}|=l$. A model produces
$f_\theta(p)=\softmax(\W\,\hidden_\theta(p))$, where $\hidden_\theta\colon
\mathcal{X}^N\to\R^h$ computes a hidden state and $\W\in\R^{l\times h}$ is the
projection, with $h\ll l$. We assume the idealized oracle that returns the full
logit vector $z=\W\,\hidden_\theta(p)\in\R^l$; the engineering needed to recover
logits from top-$K$ log-probabilities and a logit bias is treated at length
in~[\ref{ref:carlini}] and is orthogonal to what follows. We assume the final
block is a normalization (RMSNorm or LayerNorm) followed by $\W$, so the
attainable hidden states lie on a sphere $S\subseteq\R^h$ and the logits lie in
$V:=\col(\W)$.

\section{The output ideal}\label{sec:ideal}

The attainable logits are the common zeros of two families of constraints.

\paragraph{Degree-1 generators (rank).} Let $\{\nu_1,\dots,\nu_{l-h}\}$ span
$V^\perp$. Each linear form $\ell_a(z)=\langle\nu_a,z\rangle$ vanishes on every
response; these are what the SVD recovers as the directions with zero singular
value.

\paragraph{Degree-2 generator (normalization).} Because $\|\hidden\|$ is fixed
by normalization, $z=\W\hidden$ satisfies a single quadric $q(z)=z^\top\Ahat
z-1=0$, where $\Ahat$ is symmetric positive semidefinite of rank $h$ with
$\ker\Ahat=V^\perp$. With $U\in\R^{l\times h}$ an orthonormal basis of $V$ and
$x=U^\top z$, this reads $x^\top A x=1$ for a positive definite $A\in\R^{h\times
h}$, and the recovery factorization $\W=U M^{-1}O$ with $A=M^\top M$ and $O$
orthogonal follows~[\ref{ref:carlini}].

Thus the ideal is
$\mathcal{I}=\langle\ell_1,\dots,\ell_{l-h},\,q\rangle$ and the output manifold
is the ellipsoid $\mathcal{M}=\{\ell_a=0,\;q=1\}$. Two structural facts are worth
recording.

\begin{remark}[Frobenius triviality]\label{rem:frobenius}
Passing to the exterior ideal generated by differentials, the generators are the
constant covectors $\nu_a$ and the one-form $dq=2\Ahat z$. All are closed, so
$\mathcal{I}$ is a differential ideal generated by $1$-forms: the system is
Frobenius-integrable. This is the structural reason the attack is one-shot
linear algebra with no integrability obstruction, and it tells us in advance that
the heavier Cartan--K\"ahler machinery can only become necessary when more than
one layer is involved (Section~\ref{sec:below}).
\end{remark}

\begin{remark}[Architecture fingerprint]
The count of degree-1 generators distinguishes normalizations: LayerNorm centers
before normalizing, adding one linear generator and dropping the effective
dimension by one, which is the singular-value signal used
in~[\ref{ref:carlini}] to tell LayerNorm from RMSNorm.
\end{remark}

\section{Polar space, gauge, and recovery}\label{sec:polar}

In the Cartan--K\"ahler theory the \emph{polar space} of an integral element
controls which directions extend it~[\ref{ref:eds},\ref{ref:bryant}]. Here the
governing generator is the quadric, so the relevant object is the classical
polar (conjugate) space of that quadric.

\begin{proposition}[Polar space is the tangent space]\label{prop:polar}
At a point $z_0\in\mathcal{M}$ the polar space $H(z_0)=(\Ahat z_0)^\perp$ equals
the tangent space $T_{z_0}\mathcal{M}$. Consequently recovering the layer is
recovering the field of polar hyperplanes, each tangency condition
$z_0^\top\Ahat v=0$ being one linear equation in the entries of $\Ahat$.
\end{proposition}

We verify this independently: computing a tangent direction of the output
manifold by finite differences (without using the recovered $\Ahat$), its cosine
with the recovered polar normal is $5.8\times10^{-7}$, limited only by the
finite-difference step.

\paragraph{The gauge.} The symmetric form $\Ahat=M^\top M$ fixes only the Gram
part of $\W$; the antisymmetric complement, of dimension $\binom{h}{2}$, is free.
This is the orthogonal gauge: under $\W\mapsto\W O^\top$, $\hidden\mapsto
O\hidden$, the logits $z=\W O^\top O\hidden=\W\hidden$ are unchanged for every
prompt, so the entire output distribution is $O(h)$-invariant and no statistic of
the logits can distinguish points of the orbit. Recovery up to $O(h)$ is
therefore information-theoretically optimal from logits alone.

On the toy model ($h=64$), Cholesky followed by a \emph{scaled-orthogonal}
alignment reconstructs $\W$ to root-mean-square error $6\times10^{-16}$---the
same machine precision as a full affine alignment with $h^2$ free
parameters---while the aligning rotation has $\|\Omega-I\|_F\approx 11.5\approx
\sqrt{2h}$, i.e.\ a generic rotation. The quadric thus resolves the symmetric
content exactly and leaves precisely the $O(h)$ gauge, as predicted.

\section{Regularity, and what breaks it}\label{sec:reg}

Recovery is well posed under the affine analogues of K\"ahler-regularity:

\begin{description}
\item[(R1) Spectral gap.] A strictly positive multiplicative gap between
$\sigma_h$ and $\sigma_{h+1}$, equivalently full rank $h$ of both the
hidden-state matrix and $\W$.
\item[(R2) Nondegenerate quadric.] $\Ahat$ positive definite of rank $h$ on $V$;
equivalently the activations are not confined to a proper sub-variety, so the
ellipsoid-fitting system has full rank.
\item[(R3) Uniformity.] (R1)--(R2) hold on a neighborhood; automatic once
$\Ahat\succ0$.
\end{description}

These are not decorative. Table~\ref{tab:noise} sweeps i.i.d.\ logit noise (the
defense of~[\ref{ref:carlini}, App.~I]) and shows two distinct laws: the rank
gap degrades like $1/\sigma$ but stays above $1$ even at $\sigma=1$, so $\hat h$
is recovered at every level---\emph{rank is robust}; while the orthogonal-recovery
error grows linearly, $\mathrm{RMS}\approx0.036\,\sigma$---\emph{the projection is
fragile} and needs the regularity to hold tightly. Confining activations to an
effective-rank subspace (an R1 violation) makes the attack return the effective
rank rather than the nominal width---reproducing the GPT-2-Small anomaly
of~[\ref{ref:carlini}], where $757$ was recovered for a $768$-dimensional model.
Figure~\ref{fig:main} (left) plots both laws.

\begin{table}[t]
\centering
\caption{Noise sweep on the toy model ($h=64$, unit-scale logits). Rank
recovery (R1) is robust; projection recovery (R2/R3) degrades linearly.}
\label{tab:noise}
\begin{tabular}{rccc}
\toprule
$\sigma$ & $\hat h$ & gap $\sigma_h/\sigma_{h+1}$ & orthogonal RMS \\
\midrule
$10^{-4}$ & 64 & $2.48\times10^{4}$ & $3.0\times10^{-6}$ \\
$10^{-3}$ & 64 & $2.49\times10^{3}$ & $3.2\times10^{-5}$ \\
$10^{-2}$ & 64 & $2.48\times10^{2}$ & $3.0\times10^{-4}$ \\
$10^{-1}$ & 64 & $2.46\times10^{1}$ & $2.9\times10^{-3}$ \\
$1.0$     & 64 & $2.56\times10^{0}$ & $3.6\times10^{-2}$ \\
\bottomrule
\end{tabular}
\end{table}

\section{Below the last layer}\label{sec:below}

After the single-layer attack, the hidden states $\hidden(p)$ are themselves
known up to the global rotation. We now ask what this reveals about the block
beneath. Consider a toy with $k$-dimensional content fed through an MLP block
with a residual connection,
\[
  \hidden=\mathrm{norm}\bigl(x+\W_2\,\phi(\W_1 x)\bigr),\qquad x=Bs,\;s\in\R^k,
\]
with $\phi=\tanh$, hidden width $m$, residual width $h$. The attainable hidden
states now lie on a \emph{curved} $k$-dimensional submanifold of the sphere whose
linear span can be far larger than $k$. The intuition is the same as a circle in
the plane: a circle is a one-dimensional object, yet it does not fit in any
single line---its linear span is two-dimensional. A curved $k$-dimensional
manifold likewise needs more than $k$ linear dimensions to contain it, and the
nonlinearity $\phi$ is precisely what bends the content manifold so that its span
inflates above its true dimension.

\begin{figure}[t]
  \centering
  \includegraphics[width=\linewidth]{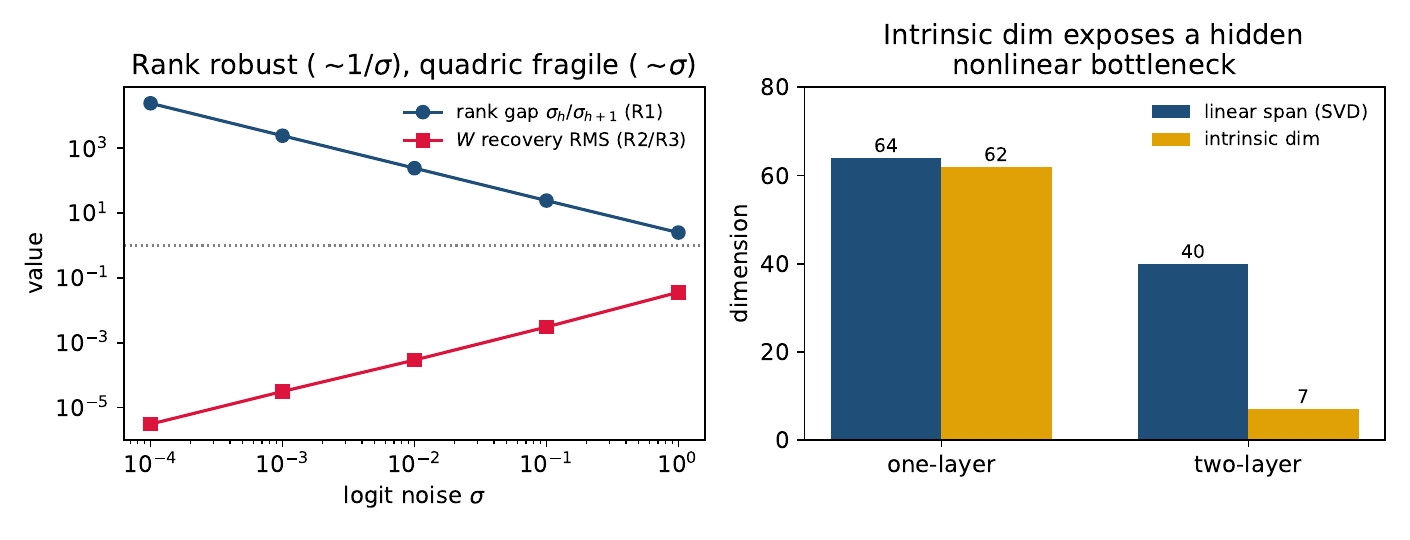}
  \caption{\textbf{Left:} regularity is load-bearing. Under logit noise the rank
  gap (R1) decays like $1/\sigma$ but never falls below $1$; the projection
  recovery error (R2/R3) grows linearly. \textbf{Right:} below the last layer,
  the linear span (what the SVD reports) overstates the content; the intrinsic
  manifold dimension recovers it and exposes the nonlinear bottleneck.}
  \label{fig:main}
\end{figure}

\paragraph{An observable the linear attack misses.} The SVD sees only the linear
span. On a two-layer toy with $h=64$, $k=8$, $m=32$ the span is $40$, the quadric
is positive definite and well posed (design rank $820/820$), and the attack
returns a clean rank-$40$ projection with no internal sign that anything is
amiss---it reports a featureless $40$-dimensional linear layer. But the
\emph{intrinsic dimension} of the recovered manifold, estimated by local PCA, is
$7$--$8$: the true content dimension. The gap between span ($40$) and intrinsic
dimension ($8$) is the fingerprint of a low-rank nonlinear bottleneck. The
one-layer control gives intrinsic dimension $62$ against span $64$, i.e.\
$\text{span}-1$ (the sphere), with no gap. In the EDS reading the intrinsic
dimension is the first Cartan character and the span--intrinsic gap is the
osculating (second-order) data the linear attack discards.
Figure~\ref{fig:main} (right) shows the contrast.

\section{The identifiability wall}\label{sec:wall}

The observable is the hidden-state manifold $\mathcal{S}$ up to the global
orthogonal gauge. We characterize the sublayer $x\mapsto x+\W_2\phi(\W_1 x)$
relative to it. The governing intuition is simple: one can only learn about parts
of a network that the inputs actually exercise, and only up to relabelings that
leave every output unchanged. Both effects turn out to be large here.

\begin{proposition}[Identifiability boundary, toy model]\label{prop:wall}
From $\mathcal{S}$ the sublayer is determined only by its input--output behaviour
on the input support $\mathcal{X}=\col(B)$, and only up to: the global rotation
$O(h)$; input reparametrization $GL(k)$ on $\mathcal{X}$; and, for minimal
$\tanh$ realizations, neuron sign-permutation~[\ref{ref:sussmann}]. The following
are free fibers, hence unrecoverable:
\begin{enumerate}
\item[\textup{(a)}] the action of $\W_1$ on $\mathcal{X}^\perp$;
\item[\textup{(b)}] the MLP width $m$ (only the minimal realization on
$\mathcal{X}$ is pinned);
\item[\textup{(c)}] $B$ beyond its column space (a full $GL(k)$);
\item[\textup{(d)}] the parametrization of the input distribution---only its
support matters.
\end{enumerate}
\end{proposition}

We make the fibers concrete with bit-level examples (all logit differences below
are at machine precision, $\sim10^{-14}$).

\begin{description}
\item[(a) Off-support freedom.] Let $\Delta=\Delta(I-P_{\mathcal{X}})$ act only on
$\mathcal{X}^\perp$. Then $\W_1$ and $\W_1+\Delta$ produce identical outputs. For
$h=64$, $k=8$, $m=32$ this hides $m(h-k)=1792$ of the $2048$ parameters of $\W_1$
($87.5\%$), with $\|\Delta\|_F\approx 42$.
\item[(b) Width.] Appending a cancelling neuron pair
($\W_1\!\to\!\begin{psmallmatrix}\W_1\\w\\w\end{psmallmatrix}$,
$\W_2\!\to\![\W_2\;c\;{-}c]$) changes the width from $m=32$ to $34$ while leaving
every output unchanged.
\item[(c) Input reparametrization.] Replacing $B$ by $BR$ for $R\in GL(k)$ leaves
the attainable input set, hence $\mathcal{S}$, unchanged.
\end{description}

The interpretation is sharp: the part of the sublayer one would most want, the
full $\W_1$, lies almost entirely inside the fiber. This is the mechanical reason
the attack stops at one layer---the observation map has a large kernel below the
last linear layer---and it reframes the open ``extend beyond one layer'' problem
of~[\ref{ref:carlini}] as an identifiability question rather than an algorithmic
one.

\section{Scope, novelty, and related work}\label{sec:scope}

We are explicit about what is and is not new.
\emph{What is reproduced:} the last-layer recovery, including recovery up to an
orthogonal matrix via the normalization-induced ellipsoid, is
from~[\ref{ref:carlini}]. \emph{What the geometric language adds:} a single
account in which rank recovery, ellipsoid recovery, the $O(h)$ gauge, and the
regularity conditions are one structure---the ideal, its polar space, and its
K\"ahler-regularity---together with the Frobenius observation
(Remark~\ref{rem:frobenius}) that explains why the single-layer attack is
closed-form. We stress that this framing is organizing, not enabling. \emph{What
is a genuinely new observation, though built on known tools:} the intrinsic
dimension of the recoverable manifold as an extraction observable
(Section~\ref{sec:below}), and the explicit identifiability boundary with its
fibers (Section~\ref{sec:wall}). The first leans on the established fact that
representations have low intrinsic dimension~[\ref{ref:ansuini}]; the second on
classical network identifiability~[\ref{ref:sussmann}]. Concurrent and prior work
on the same attack family includes~[\ref{ref:finlayson},\ref{ref:zanella}]. The
EDS toolkit we borrow from is that of~[\ref{ref:bryant}] and its Lie-algebroid
extension~[\ref{ref:eds}].

We did \emph{not} recover the sublayer parameters, and
Section~\ref{sec:wall} indicates this is blocked by identifiability rather than
by a missing technique. A clean impossibility theorem for realistic
architectures, with a matching positive recovery result on the identifiable
quotient, is the natural next target and is left open.

\section{Conclusion}

Last-layer model stealing has a tidy geometric description: the logits trace the
integral variety of an ideal with a linear and a quadratic generator, the
quadric's polar space is the manifold's tangent space and controls recovery, and
the attack succeeds up to the orthogonal gauge exactly under regularity
conditions that we showed are load-bearing. One layer down, the linear span the
SVD reports can hide a low-rank nonlinear bottleneck that the intrinsic
dimension reveals, and the sublayer's parameters live largely in an explicit
non-identifiable fiber. The geometry is a clarifying lens; the wall beneath the
last layer is real and, we argue, the more fruitful object of study.

\paragraph{Reproducibility.} Every number in this note is produced by a small
NumPy library and a single reproduction script; figures are regenerated by one
command. Code: \url{https://github.com/nssprogrammer/eds-stealing}.


\begin{thebibliography}{9}

\bibitem{ref:carlini}\label{ref:carlini}
N.~Carlini et al.,
\textit{Stealing Part of a Production Language Model.}
ICML 2024. arXiv:2403.06634.

\bibitem{ref:eds}\label{ref:eds}
S.~Hohloch, T.~Mestdag, K.~Yasaka,
\textit{The Cartan--K\"ahler theorem for exterior differential systems on
transitive Lie algebroids.} arXiv:2605.29083 (2026).

\bibitem{ref:bryant}\label{ref:bryant}
R.~L.~Bryant, S.~S.~Chern, R.~B.~Gardner, H.~L.~Goldschmidt, P.~A.~Griffiths,
\textit{Exterior Differential Systems.} Springer, 1991.

\bibitem{ref:sussmann}\label{ref:sussmann}
H.~J.~Sussmann,
\textit{Uniqueness of the weights for minimal feedforward nets with a given
input--output map.} Neural Networks 5(4):589--593, 1992.

\bibitem{ref:ansuini}\label{ref:ansuini}
A.~Ansuini, A.~Laio, J.~H.~Macke, D.~Zoccolan,
\textit{Intrinsic dimension of data representations in deep neural networks.}
NeurIPS 2019. arXiv:1905.12784.

\bibitem{ref:finlayson}\label{ref:finlayson}
M.~Finlayson, S.~Swayamdipta, X.~Ren,
\textit{Logits of API-protected LLMs leak proprietary information.}
arXiv:2403.09539 (2024).

\bibitem{ref:zanella}\label{ref:zanella}
S.~Zanella-B\'eguelin, S.~Tople, A.~Paverd, B.~K\"opf,
\textit{Grey-box extraction of natural language models.} ICML 2021.

\end{thebibliography}
\end{document}